\documentclass[11pt]{article}

\usepackage[final]{acl}

\usepackage{times}
\usepackage{latexsym}

\usepackage[T1]{fontenc}
\usepackage[utf8]{inputenc}

\usepackage{microtype}

\usepackage{inconsolata}

\usepackage{graphicx}
\usepackage{xcolor}
\usepackage{forest}
\usepackage{fontawesome5}
\usepackage{amsmath}
\usepackage[nolist]{acronym}
\usepackage[normalem]{ulem}

\definecolor{colError}{RGB}{255,210,210}
\definecolor{colUnintended}{RGB}{255, 223, 186}
\definecolor{colIntended}{RGB}{210,225,255}

\definecolor{colClear}{RGB}{200,240,210}
\definecolor{colUnclear}{RGB}{220,205,240}

\definecolor{colAvailable}{RGB}{225,245,230}
\definecolor{colUnavailable}{RGB}{225,245,230}
\definecolor{colMultiple}{RGB}{240,225,250}
\definecolor{colUnspecific}{RGB}{240,225,250}

\title{Not All Fallbacks Are Failures: Understanding and Recovering from Fallbacks in Mobile Voice Assistants}

\author{Phillip Schneider\textsuperscript{1,2}\thanks{Equal contribution.}, Alexandre Mercier\textsuperscript{2}\footnotemark[1], Joshua Oehms\textsuperscript{2}, \\ {\bf Kristiina Jokinen\textsuperscript{3}, Florian Matthes\textsuperscript{2}} \\
         $^1$ALMA PHIL, Germany \\
         $^2$Technical University of Munich, Department of Computer Science, Germany \\ 
         $^3$CNRS-AIST Joint Robotics Laboratory (JRL), Japan \\
         \texttt{\{phillip.schneider, alex.mercier, joshua.oehms, matthes\}@tum.de} \\
         \texttt{kristiina.jokinen@aist.go.jp}}

\begin{acronym}
    \acro{nlp}[NLP]{natural language processing}
    \acro{plm}[PLM]{pretrained language model}
    \acroplural{plm}[PLMs]{pretrained language models}
    \acro{ai}[AI]{artificial intelligence}
    \acro{nlg}[NLG]{Natural language generation}
    \acro{asr}[ASR]{automated speech recognition}
    \acro{auc}[AUC]{area under the curve}
    \acro{diet}[DIET]{Dual Intent and Entity Transformer}
    \acro{llm}[LLM]{large language model}
    \acroplural{llm}[LLMs]{large language models}
    \acro{nlu}[NLU]{natural language understanding}
    \acro{ood}[OOD]{out-of-domain}
    \acro{rag}[RAG]{retrieval-augmented generation}
    \acro{roc}[ROC]{receiver operating characteristic}
    \acro{sd}[SD]{standard deviation}
    \acro{stt}[STT]{speech-to-text}
    \acro{svm}[SVM]{support vector machine}
    \acro{lr}[LR]{logistic regression}
\end{acronym}

\begin{document}
\maketitle
\begin{abstract}
Robust understanding of user input is a core requirement for voice assistants deployed in real-world environments. In practice, these systems encounter heterogeneous fallback situations caused by noisy audio input, transcription errors, ambiguous requests, incomplete utterances, or unintended activations. Existing systems typically respond with generic fallback messages, which do not resolve the underlying interaction failure and can degrade user experience. We study fallback handling in a deployed smartwatch-based voice assistant for general health support in everyday environments. Our analysis is based on six months of real-world usage data from more than 500 users, yielding a dataset of 3,030 anonymized, naturally occurring fallback-triggering utterances. We contribute (1) an operational taxonomy and the annotated \textsc{VoxFallbacks} dataset of these interactions, (2) a comparative evaluation of different models within a classification pipeline under practical deployment constraints, and (3) practical lessons for designing robust and cost-efficient fallback mechanisms. Results show that lightweight embedding-based classifiers outperform larger generative models on most classification tasks while requiring substantially fewer computational resources.

\end{abstract}

\section{Introduction}

Conversational agents are increasingly embedded in mobile and assistive technologies, including smartphones, smart speakers, and wearable devices \cite{seaborn2021voice}. Unlike research prototypes, deployed systems must operate in noisy and unpredictable environments where user behavior often deviates from training data and design assumptions. Voice-based interfaces face particular challenges: background noise, accidental microphone activations \citep{schonherr2021exploring}, speech recognition errors, incomplete utterances, and requests targeting unsupported capabilities \citep{tur2014detecting}. These failure modes are routine in production yet systematically underrepresented in standard benchmarks, which typically assume well-formed, in-domain inputs \citep{larson2019evaluation}. Deployed systems must handle a large fraction of interactions that fail early in the pipeline. These failures are commonly routed to a fallback mechanism, which prevents incorrect system actions but provides limited support for interaction recovery. Standard fallback strategies often rely on generic responses such as rephrasing requests or offering help, without distinguishing between different underlying causes of failure.

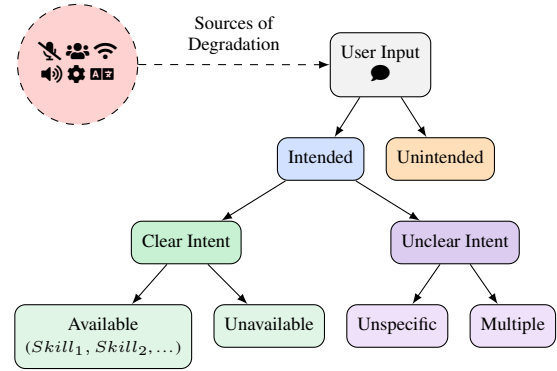
\begin{figure}[t]
\centering
\begin{forest}
for tree={
   draw,
   rounded corners,
   align=center,
   edge={-latex},
   l sep=15pt,
   s sep=8pt,
   inner sep=4pt,
   font=\scriptsize
}
% Level 1
[, phantom
% Left Node (Formerly Right Node): Dashed error sources now on the left
  [\faMicrophoneSlash \ \faUsers \ \faWifi \ \\ \faVolumeUp \ \faCog \ \faLanguage, 
   circle,              
   rounded corners=0pt, 
   draw=black,
   dashed,              
   fill=colError,
   tier=level1,         
   no edge,              
   xshift=-4cm,        % <--- Reversed to negative to shift left
    % 'node[above, midway]' puts the text right in the middle on top of the arrow:
   tikz={\draw[-latex, dashed] () -- node[above, midway, font=\scriptsize, align=center] {Sources of \\ Degradation} (userInput);}
  ]
% Right Node (Formerly Left Node): Main tree structure now on the right
  [User Input \\ \faComment, fill=gray!10, name=userInput
     [Intended, fill=colIntended
         % Level 2
         [Clear Intent, fill=colClear
             % Level 3
             [Available \\{\tiny ($Skill_1, Skill_2$, \ldots)}, fill=colAvailable]
             [Unavailable, fill=colUnavailable]
         ]
         [Unclear Intent, fill=colUnclear
             [Unspecific, fill=colUnspecific]
             [Multiple, fill=colMultiple]
         ]
     ]
     [Unintended, fill=colUnintended]
  ]
]
\end{forest}

\caption{Classification taxonomy for fallback handling.}
\label{fig:taxonomy}
\end{figure}

This work studies fallback handling in a deployed smartwatch-based voice assistant that provides health-related support in everyday environments. The system is used by older adults and individuals with chronic conditions, making robust interaction a practical requirement rather than a desirable feature. Analysis of real-world usage logs revealed that fallback-triggering utterances arise from diverse sources, including ambiguous requests, unsupported functionality, accidental activations, transcription errors, and recoverable requests that the assistant failed to interpret correctly. Rather than representing a single failure mode, these utterances correspond to different underlying interaction problems and therefore require different recovery strategies, such as clarification, capability explanations, or skill recovery. Crucially, for accidental activations, the system must accurately discern when not to react; failing to suppress such responses results in intrusive behavior that can increase user frustration and contribute to churn.

Prior work typically treats out-of-domain detection \citep{tur2014detecting}, clarification generation \citep{zhang2024clamber,sahay2025ask}, and accidental activation filtering \citep{schonherr2021exploring} as separate problems.
In voice assistants operating in real-world environments, all three appear together within the same fallback stream, demanding a unified detection and intervention approach under practical constraints of latency, computational cost, and maintainability. We therefore frame fallback handling as a structured recovery problem, focusing on selecting appropriate recovery actions rather than end-to-end recovery outcomes. Using six months of real-world usage data, we construct and analyze a dataset of anonymized fallback-triggering utterances and investigate how fallback interactions can be routed to appropriate recovery actions under practical deployment constraints.

Our contributions are threefold: we (1) provide an operational taxonomy and the annotated \textsc{VoxFallbacks} dataset\footnote{\url{https://github.com/philotron/VoxFallbacks}} with 3,030 naturally occurring fallback-triggering utterances collected from a wearable smartwatch voice assistant, (2) evaluate multiple classification approaches for routing fallback interactions to appropriate recovery strategies under deployment-oriented constraints, including accuracy, latency, and computational cost, and (3) derive practical insights showing that lightweight embedding classifiers outperform instruction-tuned \acp{llm} for fallback routing.

Overall, the paper highlights that fallback handling should not be viewed as a single error state but as a structured recovery process. Rather than defaulting to resource-intensive generative models for every component, which is a prevailing trend that warrants closer empirical scrutiny regarding actual utility, our findings demonstrate how combining taxonomy-driven classification with strictly targeted use of \acp{llm} can improve robustness in real-world voice assistants while maintaining the efficiency requirements of deployed systems.

\section{Related Work}

While fragmented across separate research streams, fallback handling comprises three main areas.

\paragraph{Accidental Activations \& Unintended Speech.} Wearable devices frequently trigger on non-directed speech, leading to intrusive system behavior. Prior research treats this primarily as a detection task, focusing on wake-word false positives \citep{larson2019evaluation, schonherr2021exploring} or analyzing linguistic breakdown signals like filled pauses and repetitions \citep{alloatti2024tag, ngo2025mmwat}. While these studies excel at filtering, they often operate in isolation from the broader conversation flow. We extend this by quantifying the prevalence of unintended speech within a production stream and demonstrating that lightweight classifiers can effectively gate these interactions before they reach more costly modules.

\paragraph{Out-of-Domain Requests \& Conversational Breakdown.} Traditional approaches to conversational breakdown focus on recovery strategies, such as re-asking, apologizing, or capability disclosure \citep{benner2021what, alghamdi2024system, stevens2025art}. Related design-science research emphasizes collaborative repair \citep{gnewuch2025overcoming} and differentiating between system-initiated and user-initiated recovery \citep{feng2023convey}. Concurrently, \ac{ood} detection has historically relied on confidence thresholds or prototype-based discriminators \citep{tur2014detecting}. We bridge these streams by refining the \ac{ood} framing: we distinguish between requests that are genuinely unsupported (requiring limitation explanations) and those that are merely underspecified or mis-transcribed (requiring re-routing).

\paragraph{Ambiguous Utterances \& Clarification.} As reliance on \acp{llm} grows, generating effective clarifications has become a central challenge. Recent findings indicate that \acp{llm} often default to silent intent assumption or over-generic probes rather than meaningful clarification \citep{zhang2024clamber, zhang2025modeling, sahay2025ask}. Architectural approaches that pair clarity classifiers with specialized generation modules \citep{murzaku2025eclair} align with our design. Our work empirically validates this decoupling: lightweight models provide robust classification, while \acp{llm} are reserved for generating contextually appropriate clarifications only when intent is confirmed as ambiguous.

\section{Dataset Construction}
\paragraph{Data Source \& Collection.}

The primary data source for this study is a smartwatch-based voice assistant tailored for elderly and chronically ill individuals across the German-speaking DACH region. Designed to support people in need of care during both daily living and critical situations, the assistant provides a suite of native capabilities: acute emergency call routing, active health tracking (blood pressure, pulse, or steps), interactive micro-dialogues (medication and hydration reminders), and general informational lookups (general knowledge questions or weather forecasts).

Audio capture and \ac{stt} processing are triggered via two modalities: (1) manual activation, where the user touches a dedicated complication on the smartwatch display, and (2) proactive activation, where the agent initiates a turn (e.g., a hydration reminder) and opens the microphone for an open-ended user response. 

The core pipeline of the deployed system relied on Microsoft Azure \ac{stt}\footnote{\url{https://ai.azure.com/catalog/models/Azure-Speech-Speech-to-text}} via its German Universal Language Model for transcription, and the Google Dialogflow agent framework\footnote{\url{https://docs.cloud.google.com/dialogflow/docs}} leveraging its proprietary \ac{nlu} for intent classification.

We collected single-turn user utterances that triggered the system's fallback mechanism (operationally defined as any input where the core intent classification confidence fell below a threshold) over a six-month period spanning 2024--2025. The core system capabilities remained stable throughout this window. Across the collection period, fallback interactions accounted for approximately 6–7\% of all assistant requests. The extracted fallback dataset is associated with more than 500 unique individuals aged predominantly between 60--90 years. To comply with data protection requirements, all personally identifiable information was removed, and only fully de-identified data were made available for academic analysis.

\paragraph{Data Selection \& Preprocessing.}

The raw dataset contained 7,857 fallback-triggering user inputs collected from a deployed voice assistant. To obtain a clean corpus for analysis, we applied a multi-phase preprocessing pipeline. First, we retained only dialogue-initiating utterances and removed mid-dialogue responses constrained by system expectations (e.g., yes/no or numeric answers). Second, exact duplicate utterances were discarded to reduce artifacts from repeated accidental activations and background noise. Third, empty transcriptions were removed. Fourth, all personally identifiable information, including names, addresses, facilities, and local businesses, was manually anonymized by replacement with generic German names and major cities. Finally, utterances were randomized to prevent reconstruction of user-specific interaction histories. The resulting corpus comprises 3,030 unique, fully anonymized fallback utterances.

Because we aim to characterize diverse fallback phenomena rather than raw production frequency, we deduplicated exact utterances to avoid overrepresenting repetitive events. To assess its impact, we classified all 7,857 records using a \ac{svm} (weighted F1 = 0.87). The resulting distribution differs only moderately from the annotated dataset (Jensen--Shannon distance = 0.119), with the largest shift being an increase in unintended input \cite{lin1991divergence}.

\paragraph{Data Annotation Protocol.}
\label{subsec:data_annotation}

To map fallback phenomena into actionable engineering decisions, we designed a hierarchical taxonomy using a rule-informed, inductive approach, combining theoretical frameworks from related work with a bottom-up review of production logs to match the actual capabilities of the deployed assistant. As illustrated in Figure~\ref{fig:taxonomy}, the taxonomy is structured as a decision tree with three distinct hierarchical levels:
(1) user intentionality separates accidental activations or ambient background noise (unintended) from purposeful interactions (intended);
(2) intent clarity evaluates whether an actionable objective can be extracted from intentional inputs, partitioning vague utterances into either ambiguous requests or multi-intent conflicts (unclear intent); and
(3) capability availability maps well-formed requests (clear intent) to either the assistant's supported feature set (available) or unsupported domains (unavailable).

Annotating fallback-triggering utterances is challenging due to ambiguous requests, speech recognition errors, and incomplete context. We therefore adopted a multi-phase annotation process involving four native German-speaking researchers. In an initial phase, a research assistant labeled all 3,030 utterances and iteratively discussed borderline cases with the research team to refine label definitions and decision rules (Table~\ref{tab:class_labels}). Subsequently, three researchers independently annotated the dataset, with a shared validation subset of 350 utterances labeled by all annotators. 
Inter-annotator agreement on this subset reached a Fleiss' Kappa
\citep{fleiss-1971-measuring} of $\kappa = 0.759$ (Table~\ref{tab:iaa}), indicating substantial
agreement \citep{landisMeasurementObserverAgreement1977}. Labels for the validation subset were determined by majority vote, with ties resolved through discussion. Finally, all disagreements between the refinement and independent annotation phases were reviewed jointly to produce the final labels.

\section{Experimental Setup}

\paragraph{Pipeline \& Models.}
We evaluate a two-stage classification pipeline aligned with the taxonomy hierarchy (Figure~\ref{fig:architecture}). Stage~1 distinguishes intended from unintended utterances. Inputs classified as unintended are suppressed and exit the pipeline. Stage~2 processes intended utterances, first determining whether intent is clear or unclear and, for clear-intent requests, whether the requested capability is available or unavailable.

We compare lightweight embedding-based classifiers and instruction-tuned \acp{llm}. For the embedding approach, we use the multilingual models \textit{multilingual-e5-large} (E5) \cite{wang2024e5} and \textit{BGE-M3} \cite{chen-etal-2024-bgem3}, each paired with \ac{lr}, \ac{svm}, and random forest classifiers. All classifiers use default hyperparameters unless stated otherwise; for \ac{lr} and \ac{svm}, preliminary experiments indicated that $C=10$ and \texttt{max\_iter=2000} yielded the most promising results, and for random forest we use \texttt{n\_estimators=200}. We additionally evaluate \textit{Gemma~4 E4B} \cite{gemma42026}, \textit{Qwen~3.5 4B} \cite{qwen3.5}, and \textit{GPT-4.1~nano}.\footnote{\url{https://developers.openai.com/api/docs/models/gpt-4.1-nano}} To reflect deployment constraints, we use small 4B variants of the open-source models. \acp{llm} are evaluated in zero-shot and few-shot settings with temperature 0 and reasoning disabled for consistency and latency reasons.

\paragraph{Prompt Design.} 
All prompts are provided in full in Appendix~\ref{sec:appendix}. We use the same prompt design principles across \ac{llm} experiments: prompts explicitly define the system role, describe the available target and fallback labels, and enforce a strict output format with ambiguity guardrails. In Stage 2, we evaluate two configurations. In the zero-shot configuration, the prompt exposes the complete set of 18 system skills and two fallback labels, each with a concise description. In the retrieval-conditioned configuration, the prompt instead provides only the top five intent candidates retrieved by a \textit{\ac{diet}} from \citet{bunk2020diet} trained on the deployed assistant’s training phrases, together with the same fallback labels. Thus, the retrieval-conditioned setup constrains the \ac{llm}’s prediction space to intents considered relevant by the upstream classifier.

We evaluate retrieval conditioning only as part of the complete end-to-end classification pipeline. We do not separately assess the \ac{diet} model’s recall or candidate-ranking quality. Accordingly, the reported results measure the effect of conditioning the \ac{llm} on retrieved candidates, rather than the retrieval component in isolation.

\begin{table}[b]
  \small
  \centering
  \begin{tabular}{lrr}
    \hline
    \textbf{Class} & \textbf{Count} & \textbf{Percentage} \\
    \hline
    unintended & 2,138 & 70.56\% \\
    intended/clear int./available & 731 & 24.13\% \\
    intended/clear int./unavailable & 103 & 3.40\% \\
    intended/unclear int./unspecific & 52 & 1.72\% \\
    intended/unclear int./multiple & 6 & 0.20\% \\

    \hline
    Total & 3,030 & 100.00\% \\
    \hline
  \end{tabular}
  \caption{Aggregated class distribution of the annotated fallback dataset.}
  \label{tab:class_distribution}
\end{table}

\begin{table*}[t]
\small
\centering
\resizebox{\textwidth}{!}{
\begin{tabular}{lccccccc}
\hline
\textbf{Model} & \textbf{Accuracy} & \textbf{Precision} & \textbf{Recall} & \textbf{F1} & \textbf{ROC AUC} & \textbf{Avg Precision} & \textbf{Latency (ms)}\\
\hline
E5+LR & $.908{\pm}.011$ & $.810{\pm}.021$ & $.900{\pm}.026$ & $\textbf{.852}{\pm} .018$ & $\textbf{.966}{\pm}.006$ & $\textbf{.931}{\pm} .011$ & $\textbf{.043}{\pm}.001$\\
BGE-M3+LR & $.901{\pm}.015$ & $.806{\pm}.035$ & $.879{\pm}.034$ & $.840{\pm}.023$ & $.960{\pm}.008$ & $.922{\pm}.012$ & $\textbf{.043}{\pm}.000$\\
E5+SVM & $\textbf{.913}{\pm}.015$ & $.884 { \pm} .042$ & $.814 { \pm} .028$ & $.847 { \pm} .026$ & $.964 { \pm} .006$ & $.928 { \pm} .009$ & $.619{\pm}.004$\\
BGE-M3+SVM & $.901 { \pm} .003$ & $.872 { \pm} .023$ & $.780 { \pm} .022$ & $.823 { \pm} .005$ & $.955 { \pm} .006$ & $.913 { \pm} .011$ & $.614{\pm}.014$\\
E5+random forest & $.909 { \pm} .012$ & $.908 { \pm} .031$ & $.768 { \pm} .019$ & $.832 { \pm} .021$ & $.957 { \pm} .010$ & $.922 { \pm} .014$ & $13.250{\pm}.031$\\
BGE-M3+random forest & $.904 { \pm} .005$ & $\textbf{.918} { \pm} .010$ & $.740 { \pm} .024$ & $.819 { \pm} .012$ & $.957 { \pm} .008$ & $.922 { \pm} .011$ & $13.265{\pm}.033$\\
Gemma 4 (zero-shot) & $.877 { \pm} .012$ & $.763 { \pm} .022$ & $.845 { \pm} .019$ & $.802 { \pm} .019$ & $.940 { \pm} .011$ & $.896 { \pm} .015$ & $242.191{\pm}3.003$\\
Gemma 4 (few-shot) & $.891 { \pm} .007$ & $.808 { \pm} .020$ & $.829 { \pm} .009$ & $.818 { \pm} .011$ & $.943 { \pm} .008$ & $.899 { \pm} .014$ & $228.888{\pm}4.418$\\
Qwen 3.5 (zero-shot) & $.635 { \pm} .019$ & $.443 { \pm} .014$ & $.929 { \pm} .012$ & $.600 { \pm} .012$ & $.902 { \pm} .013$ & $.851 { \pm} .018$ & $188.362{\pm}4.493$\\
Qwen 3.5 (few-shot) & $.448 { \pm} .013$ & $.347 { \pm} .006$ & $\textbf{.993} { \pm} .006$ & $.515 { \pm} .007$ & $.926 { \pm} .009$ & $.880 { \pm} .014$ & $183.568{\pm}2.684$\\
GPT-4.1 nano (zero-shot) & $.876 { \pm} .014$ & $.817 { \pm} .036$ & $.748 { \pm} .021$ & $.780 { \pm} .023$ & $.915 { \pm} .012$ & $.853 { \pm} .035$ & $637.476{\pm}52.588$\\
GPT-4.1 nano (few-shot) & $.859 { \pm} .011$ & $.756 { \pm} .026$ & $.774 { \pm} .016$ & $.764 { \pm} .016$ & $.912 { \pm} .014$ & $.825 { \pm} .044$ & $1660.601{\pm}286.86$\\
\hline
\end{tabular}
}
\caption{Binary classification results for the first stage of the pipeline. Bold values indicate best performance.}
\label{tab:results_binary}
\end{table*}

\paragraph{Evaluation Process.}

Models are evaluated using stratified 5-fold cross-validation. For Stage~1, we report accuracy, precision, recall, F1, ROC AUC, and average precision. For Stage~2, we report macro-averaged precision, recall, and F1 to account for class imbalance. Since ROC AUC requires a continuous confidence score, we derive probabilities for \acp{llm} from the log-probabilities of the binary output tokens. Concretely, let $\ell_1$ and $\ell_0$ denote the log-probabilities of tokens \texttt{"1"} and \texttt{"0"} at the first decoded position. The positive-class probability is then computed as:
\begin{equation}
    p_1 = \frac{e^{\ell_1}}{e^{\ell_1} + e^{\ell_0}}
\end{equation}
This approach follows established practice for deriving confidence scores from autoregressive model outputs \cite{malinin2021uncertainty, kuhn2023semantic}. In addition, we report mean classifier inference latency per utterance across all models. For embedding-based classifiers, this measurement covers only the classifier inference after embeddings have been computed; embedding computation itself takes approximately 40 ms per utterance.

\section{Results \& Discussion}
\subsection{Dataset Findings}

Our analysis focuses on the diversity of unique fallback utterances rather than their absolute operational frequency, providing a structural overview of the deduplicated dataset. As shown in Table~\ref{tab:class_distribution}, unintended activations constitute the largest portion of the annotated utterances (70.56\%). These unintended examples are notably longer and show higher length variance (averaging 48.28 characters, SD = 45.50) than both the overall dataset average (43.17 characters) and intended clear intent inputs (31.31 characters), frequently capturing longer fragments of background speech or ambient audio.

Conversely, intended requests with a clear intent targeting available skills make up 24.13\% of the unique fallback utterances in the corpus. Within this subset of potentially recoverable errors, instances are concentrated in open-ended, high-variance domains like general knowledge searches, creative writing prompts, and news lookups, whereas highly structured utility tasks such as setting timers or measuring vitals are less frequent. Finally, semantic ambiguities, consisting of unspecific or multi-intent requests, are rare, representing less than 2\% of the unique fallback utterances combined.

In addition, our dataset exposes the rich diversity of real-world fallback triggers, demonstrating that system failures are not a monolithic error state but a highly heterogeneous spectrum spanning four distinct pipeline layers (Table~\ref{tab:pipeline_fallback_taxonomy}). At the physical layer, the data captures acoustic and environmental diversity, ranging from truncated signals like microphone cut-offs to ambient background noise (e.g., radio or TV). Moving to the \ac{stt} layer, failures shift to phonetic distortions, producing wrong transcriptions of phonetically adjacent words, such as confusing ``Tassen'', the German word for cups, with ``Tasten'', the word for buttons. The \ac{nlu} layer reveals conversational and syntactic complexity, characterized by mid-utterance self-corrections where users fluidly change their mind mid-command. Finally, the context layer highlights pragmatic diversity, capturing the critical distinction between clear but out-of-scope requests (e.g., asking to play a specific song) and side-speech directed at another person, underscoring why generic fallback responses are insufficient for real-world deployment.

\begin{table*}[t]
\small
\centering
\resizebox{\textwidth}{!}{
\begin{tabular}{lcccccc}
\hline
\textbf{Model} & \textbf{Accuracy} & \textbf{Precision (Macro)} & \textbf{Recall (Macro)} & \textbf{F1 (Macro)} & \textbf{Latency (ms)} \\
\hline
E5+LR & $.823{\pm}.010$ & $.771 { \pm} .024$ & $.821 { \pm} .030$ & $.780 { \pm} .027$ & $.049 { \pm} .002$ \\
BGE-M3+LR & $\textbf{.840} { \pm} .013$ & $\textbf{.813} { \pm} .022$ & $\textbf{.834} { \pm} .027$ & $\textbf{.810} { \pm} .015$ & $\textbf{.046} { \pm} .004$ \\
Gemma 4 & $.754 { \pm} .030$ & $.749 { \pm} .077$ & $.741 { \pm} .088$ & $.727 { \pm} .079$ & $269.746 { \pm} 29.044$ \\
Gemma 4 + retrieval & $.677{\pm}.027$ & $.734{\pm}.072$ & $.692{\pm}.074$ & $.675{\pm}.060$ & $526.452{\pm}32.023$ \\

Qwen 3.5 & $.759 { \pm} .028$ & $.696 { \pm} .074$ & $.718 { \pm} .044$ & $.679 { \pm} .050$ & $216.316 { \pm} 13.878$ \\
Qwen 3.5 + retrieval & $.717{\pm}.039$ & $.773{\pm}.084$ & $.709{\pm}.045$ & $.707{\pm}.061$ & $877.442{\pm}93.198$ \\

GPT-4.1 nano & $.762{\pm}.020$ & $.744{\pm}.076$ & $.752{\pm}.082$ & $.721{\pm}.074$ & $642.821{\pm}23.831$ \\
GPT-4.1 nano + retrieval & $.732{\pm}.024$ & $.700{\pm}.069$ & $.691{\pm}.073$ & $.674{\pm}.070$ & $531.953{\pm}27.795$ \\
\hline
\end{tabular}
}
\caption{Multi-class classification results for the second stage of the pipeline. Bold values indicate best performance.}
\label{tab:results_multi}
\end{table*}

\subsection{Classification Results}

\begin{figure}[b!]
  \includegraphics[width=\columnwidth]{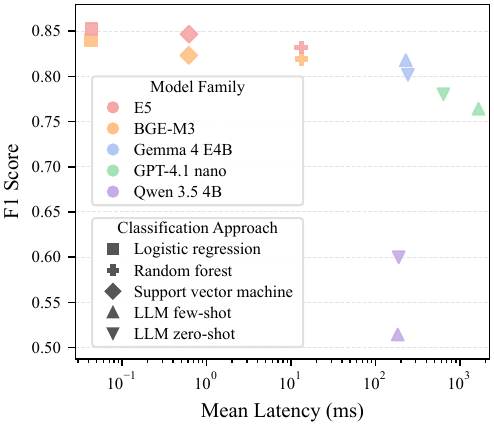}
  \caption{F1 scores of binary classification models plotted against mean latency. Colors denote model families, marker shapes denote approaches, and mean latency is shown on a logarithmic scale.}
  \label{fig:stage1_scatter}
\end{figure}

\paragraph{Binary Classification.}
Embedding-based classifiers consistently outperform the \acp{llm} in the first-stage gating of unintended versus intended inputs, both in classification quality and inference latency (Table~\ref{tab:results_binary} and Figure~\ref{fig:stage1_scatter}). Across all three classifier types (\ac{lr}, \ac{svm}, and random forest), the E5 embedding yields a higher F1 score than its BGE-M3 counterpart. The best overall result is achieved by E5\,+\,\ac{lr} (F1 = 0.852, accuracy = 0.908). More broadly, all six embedding-based models achieve F1 scores at or above the strongest \ac{llm} result (Gemma~4 few-shot, F1 = 0.818), while their classifier inference latency remains below 14 ms, compared with 184--1,661 ms for the \acp{llm}. The API-based GPT-4.1 nano configuration reaches 637 ms in the zero-shot and 1,661 ms in the few-shot setting, with these values reflecting external API communication and service-side processing rather than intrinsic model inference time. Figure~\ref{fig:stage1_scatter} makes this accuracy–latency trade-off explicit: the embedding-based models occupy the favorable region of the plot, combining higher F1 with substantially lower latency than the generative models. Hence, for this deployment setting, embedding-based classifiers are the preferred choice for first-stage gating, particularly when labeled training data are available.

Furthermore, Qwen~3.5 exhibits a typical generative failure mode: it achieves very high recall on the intended class (0.929/0.993) but suffers from collapsing precision (0.443/0.347), labeling almost everything as intended and consequently failing to suppress accidental activations. Few-shot prompting exacerbates this issue, reducing accuracy to 0.448, below the majority baseline. More broadly, few-shot prompting is not consistently beneficial for binary classification: Gemma~4 improves slightly (0.802 to 0.818) F1, whereas GPT-4.1~nano degrades (0.780 to 0.764). These results suggest that adding demonstrations can introduce decision bias without providing a useful signal for reliably improving the \ac{llm}'s ability to distinguish intended from unintended speech.

\paragraph{Multi-class Classification.}
Skill routing in the second stage is a single-label multi-class classification task that is fundamentally more complex than binary filtering, requiring semantic disambiguation across 18 skills and two fallback labels (Table~\ref{tab:results_multi}). Specifically, the two unclear labels (``multiple'' and ``unspecific'') are combined into a single ``unclear'' label. A similar picture emerges as in the first stage: the embedding-based classifiers achieve higher F1 scores than the \acp{llm} while operating at substantially lower latency, with the F1 gap being even larger in the multi-class setting. BGE-M3\,+\,\ac{lr} performs best overall (macro F1 = 0.810, accuracy = 0.840), while the best-performing \ac{llm}, Gemma~4, reaches only a macro F1 of 0.727. The embedding-based classifiers also have substantially lower classifier inference latency ($<$0.05\,ms vs.\ 216--877\,ms), although embedding computation adds approximately 40 ms per utterance. 

Notably, the relative performance of the two embedding models depends on the classification task. Whereas E5 consistently outperforms BGE-M3 across all classifier types in the binary classification setting, BGE-M3 combined with logistic regression achieves the highest macro F1 in the multi-class task. This suggests that embedding models should be evaluated separately for the specific classification tasks in which they are deployed rather than assuming that performance on one task generalizes to another.

Retrieval-conditioned prompting does not provide a consistent performance improvement across the three \acp{llm}. Its effect differs substantially by model. For Gemma~4, retrieval reduces macro F1 from 0.727 to 0.675. GPT-4.1~nano shows a similar pattern, with macro F1 decreasing from 0.721 to 0.674. Qwen~3.5 behaves differently: retrieval increases macro precision from 0.696 to 0.773 and macro F1 from 0.679 to 0.707, while recall decreases slightly from 0.718 to 0.709 and accuracy from 0.759 to 0.717. Thus, retrieval can alter the precision--recall trade-off, but its benefit is model-dependent rather than universally positive.

One possible explanation is that restricting the \ac{llm} to a small set of retrieved candidates reduces the number of semantically plausible skills considered during prediction, which may suppress some incorrect skill assignments but can also prevent the model from selecting the correct skill when it is absent from the retrieved set. Because we do not evaluate the retrieval component independently, we cannot determine from these results whether such candidate omissions are the primary cause of the observed degradation. The results therefore support interpreting retrieval as a strong conditioning mechanism that changes the \ac{llm}'s decision space, rather than as an independently validated improvement to the routing system. The additional retrieval step also increases latency substantially for Gemma~4 and Qwen~3.5.

\subsection{Failure Modes} 
For our deployed voice assistant, error costs are asymmetric, meaning their distribution is more significant than their total rate. Misrouting between available skills (routing to the wrong skill) and false activations on unintended input (routing out-of-scope to an available skill) are high risk, as they trigger incorrect or intrusive actions that can reduce user trust or lead to churn. Conversely, routing a recoverable request to an unclear or out-of-domain label is low risk, as it only necessitates a clarification turn. Therefore, fallback handling should be optimized for interaction safety under asymmetric costs rather than symmetric accuracy.

Furthermore, confusions are systematic rather than random, clustering around semantically adjacent skills like daily news versus knowledge search. \acp{llm} also shift failure directions across stages: after generating excessive false positives in the binary stage, they pivot during skill routing to funnel a large share of ambiguous interactions into out-of-scope and unspecific fallback labels. Because these confusions align with the underlying structure, many can be resolved through prompt and label design without retraining. This advantage is unique to \acp{llm}, as their target labels can be modified using natural language and they adapt easily to new categories and interaction types.

\subsection{Practical Implications}

Our results offer critical lessons for deploying robust fallback handling in resource-constrained mobile voice assistants, favoring a hybrid design that prioritizes latency and cost over minor classification gains. First, fallback inputs are highly heterogeneous, mixing accidental activations, \ac{stt} errors, and \ac{ood} commands, meaning they must be filtered through a structured taxonomy rather than handled uniformly. 

Second, lightweight embedding classifiers are superior to \acp{llm} for routing because they deliver equivalent or better performance at lower latency and compute cost, which is vital for the initial filtering of massive amounts of unintended input. Our preliminary experiments suggest that \acp{llm} can generate high-quality clarification requests. However, as this was not systematically evaluated in this study, we consider \ac{llm}-based clarification generation a promising direction for future work. 

Ultimately, these findings favor a hybrid architecture that optimizes efficiency and performance. Shifting the bulk of classification onto these embedding models significantly reduces compute and serving costs. Crucially for industry deployments, this approach eliminates reliance on external \ac{llm} APIs, making local self-hosting highly feasible, whereas serving \acp{llm} requires a massive amount of resources.

\section{Conclusion}
We studied fallback handling in a deployed voice assistant using 3,030 annotated fallback utterances. We introduced the \textsc{VoxFallbacks} dataset and a three-level taxonomy that frames fallbacks as a structured recovery process. Across a two-stage routing pipeline, lightweight embedding-based classifiers consistently outperform \acp{llm} for intent routing while operating at substantially lower latency. These results highlight a hybrid design that prioritizes low-latency discriminative models for routing, with generative models as a promising option for clarification, while accounting for real-world latency, cost, and privacy constraints. Future work should extend this framework to multi-turn recovery modeling with richer dialogue state tracking. It should also evaluate end-to-end task success under deployed conditions, including \ac{llm}-based clarification and stronger grounding through retrieval-conditioned classification and generation.

\section*{Limitations}
There are several limitations to both the dataset and the proposed fallback handling approach. First, the dataset consists of single-turn utterances only. This restricts the analysis to isolated interaction fragments and does not capture multi-turn recovery dynamics, which are central to real-world conversational repair. As a result, the observed failure modes and the proposed taxonomy reflect turn-level breakdowns rather than complete dialogues.

Second, the data were collected from a single deployed system in a specific application context, language, and user population. The definition of a fallback is therefore tied to one \ac{stt}-\ac{nlu} pipeline and its confidence thresholding behavior. While this makes the dataset operationally grounded, it also means that the resulting categories should be interpreted as system-specific failure patterns rather than a universal taxonomy of dialogue breakdowns.

A further limitation lies in how the data were labeled. Annotations were based on transcripts without the full conversation history, and they were completed by researchers rather than the users themselves. Consequently, conclusions about what the users actually intended are just educated guesses rather than the ground truth.

On the modeling side, our evaluation is restricted to lightweight, small \acp{llm} and embedding-based classifiers, motivated by latency, cost, and privacy constraints. While this reflects realistic deployment conditions, it limits conclusions about larger or more capable models. Additionally, evaluating \ac{llm}-based clarification generation remains an open challenge, as we do not measure end-to-end task success or whether clarifications actually resolve user goals during interaction. Finally, while our discussion addresses the asymmetric costs of safety-critical errors (such as highly damaging false negatives like missed emergency intents), our performance evaluation treats all error types uniformly.

\section*{Ethics Statement}

This work complies with relevant data protection and research ethics standards. All data used in this study were collected and processed in accordance with GDPR (General Data Protection Regulation) regulations. To ensure privacy protection, the dataset intended for publication was carefully anonymized. Each utterance was manually reviewed during a multi-phase cleaning process, including independent inspection by three annotators, to remove or modify any personally identifiable information. In addition, all entries were carefully screened to ensure the absence of offensive or harmful content before inclusion in the final dataset. For \ac{llm} inference using external APIs, only anonymized utterances were submitted, with no directly identifying information included.

All embedding-based experiments and open-source \ac{llm} experiments were conducted using locally available computational resources on a MacBook Pro with an Apple M3 Pro chip and 18 GB of unified memory, ensuring that no large-scale external compute infrastructure was required. Beyond data privacy, additional ethical considerations include the deployment context of conversational systems in assistive settings, where misclassification or inappropriate fallback handling may affect user experience. The proposed methods are designed to improve robustness and reduce intrusive or unhelpful system behavior, particularly in sensitive application domains such as assistive voice interaction.

\section*{Acknowledgments}
This work has been supported by funds from the Bavarian Ministry of Economic Affairs, Regional Development and
Energy (StMWi) as part of the program “BayVFP Förderlinie Digitalisierung –
Informations- und Kommunikationstechnologie”. We would like to thank David Uhlenbrock for his valuable work on the initial annotation and refinement of the dataset.

% Bibliography entries for the entire Anthology, followed by custom entries
%\bibliography{anthology,custom}
% Custom bibliography entries only
\bibliography{custom}

\appendix
\section{Appendix}
\label{sec:appendix}

\begin{figure*}[t]
  \centering
  \includegraphics[width=\textwidth]{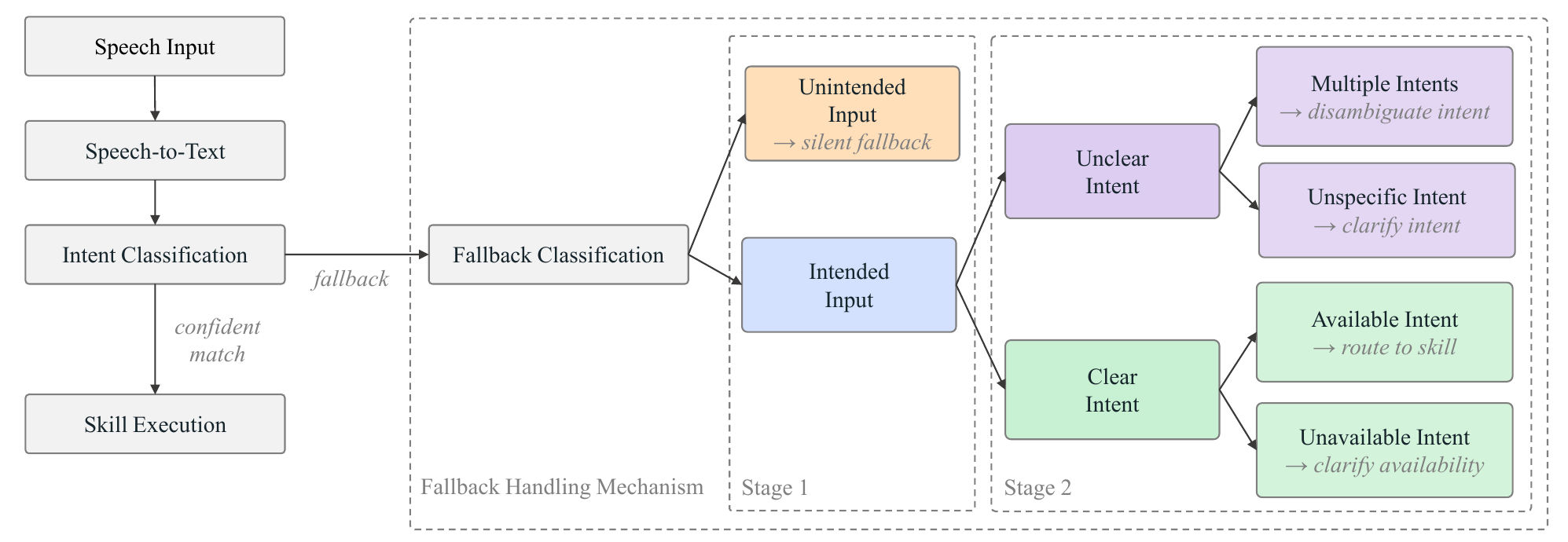}
  \caption{Hybrid fallback recovery pipeline. The intent-based dialogue manager handles matched requests; low-confidence inputs trigger a lightweight repair component consisting of a binary classification (Stage 1) and a multi-class classification for intended utterances (Stage 2). Depending on the type of the utterance, different recovery strategies are adopted.}
  \label{fig:architecture}
\end{figure*}

The Appendix provides a hybrid fallback recovery pipeline diagram (Figure~\ref{fig:architecture}), inter-annotator agreement metrics (Table~\ref{tab:iaa}), a categorization of observed interaction failures across the pipeline (Table~\ref{tab:pipeline_fallback_taxonomy}), an overview of class labels from the annotated dataset (Table~\ref{tab:class_labels}), and finally, several figures detailing the prompts used for zero-shot and few-shot classification in the first and second stages of the pipeline (Figures~\ref{fig:prompt_stage1_zero}, \ref{fig:prompt_stage1_few}, \ref{fig:prompt_stage2_zero}, and \ref{fig:prompt_stage2_augmented}).

\begin{table*}[h]
\centering
\small
\begin{tabular}{llc}
\hline
\textbf{Metric} & \textbf{Annotators} & $\boldsymbol{\kappa}$ \\
\hline
Fleiss' $\kappa$          & All (1, 2, 3) & 0.7591 \\
\hline
                           & 1 vs.\ 2      & 0.7619 \\
Pairwise Cohen's $\kappa$ & 1 vs.\ 3      & 0.7471 \\
                           & 2 vs.\ 3      & 0.7685 \\
\hline
\end{tabular}
\caption{Inter-annotator agreement among three annotators, based on 350 items and 21 categories.
Extrapolated to the whole dataset with Bootstrap 95\% CI (10{,}000 resamples) yields $\kappa = 0.7591 \pm 0.0532$
$[0.7034,\ 0.8099]$. According to \citet{landisMeasurementObserverAgreement1977}, this indicates substantial agreement among annotators.}
\label{tab:iaa}
\end{table*}

\begin{table*}[ht]
  \centering
  \small
  \begin{tabular}{p{2.5cm} p{3.4cm} p{8.6cm}}
    \hline
    \textbf{Pipeline Layer} & \textbf{Failure Category} & \textbf{Selected Dataset Examples} \\
    \hline
    Physical Layer & Microphone cut-offs &  [Welche] Neue Nachrichten gibt es \\ & & Erklär mir deine [Funktionen] \\ 
    \cline{2-3}
    (Acoustic / Signal) & Ambient noise & \ldots Packungsbeilage und fragen Sie Ihren Arzt oder Ihre Apotheke \\
    & & \ldots das Gespräch sei äußerst produktiv gewesen erzählt Trump\\
    \hline
    STT Layer & Wrong transcription & Habe ich 2 \sout{Tasten} [Tassen] Kaffee getrunken \\
    (Phonetic / Lexical) & & Bitte nicht \sout{schwören} [stören] \\
     & & \sout{Dezimale} [Maximale] Lautstärke \\
              \cline{2-3}
              & Cross-lingual homophones & \dots und dann \sout{smooth} [schmus] ich noch mal mit dir, ja. \\
              \cline{2-3}
              & Out-of-vocabulary words &  \sout{Pultmessen} [Puls messen]  \\
              & & \sout{Aufhorn} [Aufhören] \\
              && \sout{Tagessau} [Tagesschau]\\
              \cline{2-3}
              & Profanity masking & Das ist natürlich jetzt *******. \\
    \hline
    NLU Layer & Self-correction & Notfall eh nicht per Sprache machen, sondern erzähle mir einen Witz. \\
    (Semantic) & & \ldots Nein nicht den Notfall einfach sagen das Ruf Jürgen an\\
    \hline
    Context Layer & Non-addressed speech & Bis wieviel Uhr kann ich anrufen? \\
    \cline{2-3}
    (Pragmatic) & Multi-speaker dialog & Sprich bitte, du musst die Frage stellen. Ja, ich muss das erstmal lesen. Ach so lange drücken Notfall. \\
    \cline{2-3}
                  & Out-of-scope requests & Assistent Spiel das Lied vom Roger Whittaker \\
    \hline
  \end{tabular}
  \caption{Overview of observed interaction failures across the voice assistant processing pipeline, populated with real-world dataset examples. Text enclosed in square brackets indicates author-provided corrections or reconstructions of the original automatic transcriptions, including inserted or corrected words.}
  \label{tab:pipeline_fallback_taxonomy}
\end{table*}

\begin{table*}[t]
  \small
  \centering
  \begin{tabular}{p{0.43\linewidth}p{0.46\linewidth}}
    \hline
    \textbf{Class} & \textbf{Short Description} \\
    \hline

    intended.clear\_intent.available\_skill.active\_listener 
    & Starting chitchat conversation without a task-oriented goal \\

    intended.clear\_intent.available\_skill.blood\_pressure\_log 
    & Logging blood pressure measurements \\

    intended.clear\_intent.available\_skill.calling 
    & Initiating phone calls \\

    intended.clear\_intent.available\_skill.change\_volume 
    & Adjusting system sound volume \\

    intended.clear\_intent.available\_skill.emergency 
    & Triggering emergency calls \\

    intended.clear\_intent.available\_skill.general\_reminder 
    & Creating general reminders \\

    intended.clear\_intent.available\_skill.hydro 
    & Logging fluid intake \\

    intended.clear\_intent.available\_skill.knowledge\_search 
    & General knowledge question or factual information request \\

    intended.clear\_intent.available\_skill.messaging 
    & Sending messages \\

    intended.clear\_intent.available\_skill.news\_search 
    & Retrieving news-related information \\

    intended.clear\_intent.available\_skill.notebook 
    & Request to create a personal note \\

    intended.clear\_intent.available\_skill.step\_counter 
    & Request related to step count \\

    intended.clear\_intent.available\_skill.tell\_time 
    & Asking for current time \\

    intended.clear\_intent.available\_skill.time\_reminder 
    & Scheduling time-based reminders \\

    intended.clear\_intent.available\_skill.vital\_measure 
    & Request to measure vital signs such as heart frequency\\

    intended.clear\_intent.available\_skill.weather 
    & Request for weather information or forecast \\

    intended.clear\_intent.available\_skill.word\_riddle 
    & Request to play a word riddle game  \\

    intended.clear\_intent.available\_skill.writing 
    & Request for a joke, story, poem, or other generated text \\

    intended.clear\_intent.unavailable\_skill 
    & Valid intent, but unsupported by the system capabilities \\

    intended.unclear\_intent.multiple 
    & Utterance contains multiple competing intents \\

    intended.unclear\_intent.unspecific 
    & Intent is too vague or incomplete \\

    unintended 
    & Accidental activation or non-directed/background speech \\

    \hline
  \end{tabular}
  \caption{Class labels of fallback-triggering utterances with short descriptions.}
  \label{tab:class_labels}
\end{table*}

\begin{figure*}[h]
\centering
\begin{tabular}{|l|}
\hline
\begin{minipage}{0.8\textwidth}
\vspace{2mm} % Adds a little padding at the top
\footnotesize % Makes the text smaller

\begin{verbatim}
You are a classifier for a German voice assistant.

Task:
Classify whether the user utterance is clearly intended for the assistant.

Labels:
1 = intended (direct command, request, or question directed at the assistant)
0 = unintended (background speech, self-talk, conversation, filler, or noise)

Output:
Return exactly one label (0 or 1). If unsure, return 0.
\end{verbatim}

\vspace{2mm} % Adds a little padding at the bottom
\end{minipage} \\
\hline
\end{tabular}
\caption{Zero-shot prompt for binary classification in the first stage.}
\label{fig:prompt_stage1_zero}
\end{figure*}

\begin{figure*}[h]
\centering
\begin{tabular}{|l|}
\hline
\begin{minipage}{0.8\textwidth}
\vspace{2mm} % Adds a little padding at the top
\footnotesize % Makes the text smaller

\begin{verbatim}
You are a classifier for a German voice assistant.

Task:
Classify whether the user utterance is clearly intended for the assistant.

Labels:
1 = intended (direct command, request, or question directed at the assistant)
0 = unintended (background speech, self-talk, conversation, filler, or noise)

Examples:
1:
- Leiser.
- Regnet es morgen in München?
- Wie weit habe ich mein Trinkziel schon erreicht?
0:
- Tochter.
- Hast du auch so eine.
- Machst du das ja ich bin sogar bei.
- Die erste kommerzielle Landung auf dem Himmelskörper.
- Innen ist der Zug nach Europa eigentlich schon ab nein.

Output:
Return exactly one label (0 or 1). If unsure, return 0.
\end{verbatim}

\vspace{2mm} % Adds a little padding at the bottom
\end{minipage} \\
\hline
\end{tabular}
\caption{Few-shot prompt for binary classification in the first stage.}
\label{fig:prompt_stage1_few}
\end{figure*}

\begin{figure*}[h]
\centering
\begin{tabular}{|l|}
\hline
\begin{minipage}{0.8\textwidth}
\vspace{2mm} % Adds a little padding at the top
\footnotesize % Makes the text smaller

\begin{verbatim}
You are a classifier for a German voice assistant.

Task:
Classify the user utterance into one matching skill label, 
or use a fallback label if no skill label applies.

Skill labels:
- active_listener: Aktiver Zuhörer / Smalltalk
- blood_pressure_log: Blutdruckmessung protokollieren
- calling: Jemanden anrufen
- change_volume: Lautstärke ändern (lauter / leiser)
- emergency: Notfall auslösen
- general_reminder: Allgemeine Erinnerung stellen
- hydro: Trinkprotokoll
- knowledge_search: Wissensfrage beantworten
- messaging: Nachricht senden
- news_search: Aktuelle Nachrichten abfragen
- notebook: Notiz aufnehmen
- step_counter: Schrittzähler abfragen
- tell_time: Uhrzeit nennen
- time_reminder: Zeitbasierte Erinnerung / Wecker stellen
- vital_measure: Vitalzeichen messen (Blutdruck / Sauerstoff)
- weather: Wetter abfragen
- word_riddle: Worträtsel (alte Schätzchen)
- writing: Witz oder kreativen Text erzählen

Fallback labels:
- out_of_scope (clear user request, but no matching skill is available)
- unclear (intent is ambiguous, incomplete, contradictory, or too noisy to classify)

Output:
Return exactly one label and nothing else. If unsure, return 'unclear'.
\end{verbatim}

\vspace{2mm} % Adds a little padding at the bottom
\end{minipage} \\
\hline
\end{tabular}
\caption{Zero-shot prompt for multi-class classification in the second stage.}
\label{fig:prompt_stage2_zero}
\end{figure*}

\begin{figure*}[h]
\centering
\begin{tabular}{|l|}
\hline
\begin{minipage}{0.8\textwidth}
\vspace{2mm} % Adds a little padding at the top
\footnotesize % Makes the text smaller

\begin{verbatim}
You are a classifier for a German voice assistant.

Task:
Classify the user utterance into one matching skill label, 
or use a fallback label if no skill label applies.

Skill labels:
{TOP_RETRIEVED_INTENTS}

Fallback labels:
- out_of_scope (clear user request, but no matching skill is available)
- unclear (intent is ambiguous, incomplete, contradictory, or too noisy to classify)

Output:
Return exactly one label and nothing else. If unsure, return 'unclear'.
\end{verbatim}

\vspace{2mm} % Adds a little padding at the bottom
\end{minipage} \\
\hline
\end{tabular}
\caption{Retrieval-augmented prompt for multi-class classification in the second stage.}
\label{fig:prompt_stage2_augmented}
\end{figure*}

\end{document}